\relax
\documentclass[letterpaper]{article} 
\usepackage{aaai20}  
\usepackage{times}  
\usepackage{helvet} 
\usepackage{courier}  
\usepackage[hyphens]{url}  
\usepackage{graphicx} 
\usepackage{graphicx}  
\usepackage{amsfonts}       

\usepackage{algorithm}
\usepackage{algorithmic}
\usepackage{amsmath}
\usepackage{booktabs} 
\newcommand{\reals}{{\mathbb{R}}}

\newcommand{\vu}{{\bf u}}

\newcommand{\vv}{{\bf v}}

\title{Deep Kernel Learning via Random Fourier Features}

\author{%
  Jiaxuan Xie, Fanghui Liu, Kaijie Wang, Xiaolin Huang\\
  Department of Automation\\
  Shanghai Jiao Tong University\\
  Shanghai, China 200240 \\
  \texttt{\{knxie,lfhsgre\}@outlook.com, \{xiaolinhuang,kaijie\_wang\}@sjtu.edu.cn} 
}

\begin{document}

\maketitle
\begin{abstract}
Kernel learning methods are among the most effective learning methods and have been vigorously studied in the past decades. However, when tackling with complicated tasks, classical kernel methods are not
 flexible or ``rich'' enough to describe the data and hence could not yield satisfactory performance. In this paper, via Random Fourier Features (RFF), we successfully incorporate the deep architecture into kernel learning, which significantly boosts the flexibility and richness of kernel machines while keeps kernels' advantage of pairwise handling small data. With RFF, we could establish a deep structure and make every kernel in RFF layers could be trained end-to-end. Since RFF with different distributions could represent different kernels, our model has the capability of finding suitable kernels for each layer, which is much more flexible than traditional kernel-based methods where the kernel is pre-selected.
 This fact also helps yield a more sophisticated kernel cascade connection in the architecture.
 On small datasets (less than 1000 samples), for which deep learning is generally not suitable due to overfitting, our method achieves superior performance compared to advanced kernel methods. On large-scale datasets, including non-image and image classification tasks, our method also has competitive performance.
\end{abstract}

\section{Introduction}

Vast quantities of accessible and available labeled data have engendered amazing breakthroughs of deep learning methods on various tasks, such as image classification, speech recognition, machine translation and object tracking. However, when tackling problems where
labeled data are insufficient, existing deep approaches could not yield a satisfactory performance due to severe overfitting. To overcome this troublesome issue,  like meta-learning and one-shot learning, there have been some useful learning methods proposed.

Another considerable category, kernel methods,
as an important machine learning tool with solid theoretical analysis, are fully developed to cope with small data problems and have been successfully applied in many fields. Nevertheless, the shallow structure limits the learning capability of kernel methods.
Besides, standard kernel methods hypothesize the user-defined kernel effective enough to describe complex non-linear relationships, which is a weakness if we do not know a good data representation in advance.

Motivated by recent progress of deep learning, people realize that making kernel deeper could be a promising way to improve the flexibility of kernel methods \cite{cho2009kernel}, \cite{wilson2016stochastic}
.
But there are two challenges for constructing a deep kernel structure. First, a layer that calculates a kernel will map inputs in $\reals^d$ into $\reals^{m \times m}$, where $m$ is the number of samples. Pairwise considering samples is a merit of kernel methods but when $m$ is large, it becomes challenging to calculate on a big matrix, not to mention to train those matrices in multi-layers.
Second,
it is difficult to train the kernel used in each layer. In fact, subject to the traditional rule of kernel methods, kernels utilized in each layer
are still pre-given and characterized by one or several parameters. Thus, even in shallow model, it is more prevalent to tune the kernel parameters rather than to train them. Let alone the choice of different kernels, which is usually based on model selection methods but it is not suitable for dealing with the case of multiple layers.

To bridge the first challenge, we turn to the well-known kernel trick, i.e., for a kernel $k$ that satisfies the Mercer's condition, there is
\begin{align}
k(\vu,\vv)=\langle\phi(\vu),\phi(\vv)\rangle
\end{align}
where $\phi: \mathcal{X}\mapsto \mathcal{H}$ is a feature map which projects samples of input space $\mathcal{X}$ into a reproducing kernel Hilbert space $\mathcal{H}$. By applying kernel trick, one can successfully controls the computation complexity in deep kernel structure. Meanwhile, the composition of nonlinear mapping essentially could yield a more sophisticated and complex kernel as the composition of nonlinear kernels, as illustrated by \cite{cho2009kernel}. Take the decompositions of two RBF kernels as an example, the composition becomes:
\begin{equation}
\begin{split}
k^{(2)}(\mathbf{u},\mathbf{v})&=\phi^{(2)}(\phi^{(1)}(\mathbf{u}))\cdot\phi^{(2)}(\phi^{(1)}(\mathbf{v}))\\
&=
e^{-2\lambda}\textrm{exp}(-2\lambda k(\mathbf{u},\mathbf{v})).
\end{split}
\end{equation}
Although the feature mappings $\phi^{(1)}, \phi^{(2)}$ that corresponding to the features of the inner and outer kernels respectively cannot be explicitly given, the above formulation, via learning feature expressions, shows the possibility of learning approaches composed of multiple kernel layers.


This idea of using features rather than kernels also appears in large-scale kernel learning, where the calculation is kept in feature space for computational efficiency. In that field, random Fourier features (RFF) become popular and promising recently \cite{rahimi2008random}, \cite{yang2012nystrom}, \cite{sinha2016learning}. Generally, a positive semi-definite and shift-invariant kernel, namely $k$, could be represented as
\begin{equation}
\begin{split}
k(\vu,\vv)=k(\vu-\vv) &= \mathbb{E}_{\omega \sim \rho(k)}[\zeta_{\omega}(\vu)\zeta_{\omega}(\vv)^*] \\ &\mathrm{~with~} \zeta_{\omega}(\vu) = e^{j\omega^\top \vu},
\end{split}
\end{equation}
where $\rho(k)$ is a distribution associated with the kernel $k$. In practice, one could randomly draw $\omega$ from $\rho(k)$ to construct the feature map to approach the original kernel. The number of features determines the approximation accuracy, since with more $\omega$, the empirical sum fits the expectation better. RFF are used mostly for speeding up the kernel machines, which still rely on a single, given kernel. Unfortunately, there has so far been no
satisfactory approaches to tackle the latter challenge.

In this paper, we propose to use RFF in each layer to construct a deep kernel method and name it random Fourier features neural networks (RFFNet). On the one hand, the deep architecture
remarkably improves the learning ability; on the other hand, each layer of RFFNet could well approximate a kernel, from which it follows good generalization capability, perfectly suitable when there are limited available data. For different kernels, the distribution $\rho(k)$ and $\omega$ drawn from it are different. We adopt back propagation to train the distribution, which equips RFFNet the capability of finding suitable kernels for different layers. One remarkable attempt to deep kernel structure is made by \cite{mairal2016end}, which designs a formulation of feature functions to represent user-given kernels. Via training the feature function, the kernel is tuned. But since the approximation is only valid for a given type of kernels, it could be difficult to extend to other categories of kernels.
By contrast, by changing the distributions, RFF could traverse all positive semi-definite and shift-invariant kernels.

In the rest, we will introduce random Fourier features (Section 2) and construct RFFNet with detailed structures, blocks, regularization terms, and the training method (Section 3.1, 3.2). With analysis (Section 3.3) and experiments (Section 4), we show that
\begin{itemize}
\item RFFNet is an efficient structure for deep kernel learning, which significantly improves the performance of shallow kernel methods. For example, the accuracy on ``monks1" could be improved from $81.5\%$ (by SVM with RBF kernel) to $100\%$ .

\item RFFNet is suitable to learn from small data,  which is inherited from kernel learning. For example, when the number of training data is less than 1000, many deep methods are hardly used but RFFNet achieves very promising performance.

\item RFFNet is able to learn from both image and non-image data, since RFF cover a various of kernels. For example, on non-image data ``EEG", accuracy of RFFNet is $98.1\%$ and that for MLP is $92.5\%$; on image data ``CIFAR-10'', accuracy of RFFNet is $84.6\%$ (with no data argumentation), comparable to $86.0\%$ archived by VGG-16.
\end{itemize}

\noindent\textbf{Related advanced kernel learning methods:} Since the proposal of deep neural networks, people have tried to enhance the flexibility of kernel methods in order to deal with complicated data.
With deeper architectures, deep kernel learning (DKL) is expected to have better flexibility and the richness of representations is expected to get dramatic boosting from shallow kernel-based methods.
In \cite{cho2009kernel}, a new family of arc-cosine kernels are proposed via the integral representation to construct a multilayer kernel machine. However, this methodology is present for the
specialized kernel and could not be applied to other common kernels. There also exist some works that are related
to the Bayesian methodology.  For instance, Smola et al. \cite{scholkopf2001learning} combine Bayesian estimation framework with kernel methods and
Wilson et al. \cite{wilson2016stochastic}, \cite{wilson2016deep}  try to learn kernels through the marginal likelihood of a Gaussian process, but these methods usually require an extra feature extraction module such as the MLP for vectors or the deep network for images.

\section{Random Fourier Features}
Let us first introduce the Random Fourier Feature method (RFF), the foundation of the proposed model. RFF originates from harmonic analysis \cite{rahimi2008random} and has become a powerful tool to approximate the feature mapping of a kernel. Here, the kernel should  satisfy the following two conditions: i) the kernel should be shift-invariant, i.e., $k(\mathbf{u},\mathbf{v})=k(\mathbf{u}-\mathbf{v})$; ii) the kernel should be positive semi-definite. The two conditions together guarantee that $k(\mathbf{z})$, where $\mathbf{z}=\mathbf{u}-\mathbf{v}$, is the Fourier transform of a measure. The Bochner's theorem provides the key insight
behind this transformation:
\newtheorem{theo}{Theorem}
\begin{theo}
(Bochner \cite{rudin1962fourier}). A continuous kernel $k(\mathbf{u},\mathbf{v})=k(\mathbf{u}-\mathbf{v})$ on $\reals^d$ is positive definite if and only if k($\delta$) is the
Fourier transform of a non-negative measure.
\end{theo}
Accordingly, with $\zeta_{\omega}(\mathbf{u})=$  $e^{j\omega^\top\mathbf{u}}$, $k(\mathbf{u}-\mathbf{v})$ can be regarded as the following expectation over a distribution $p$, i.e.,
\begin{align}\label{rff}
k(\mathbf{u}-\mathbf{v})&=\int_{\reals^d}p(\omega)e^{j\omega^\top(\mathbf{u}-\mathbf{v})}d\omega\\&=
\mathbb{E}
_{\omega}[\zeta_{\omega}(\mathbf{u})\zeta_{\omega}(\mathbf{v})^*],
\end{align}
In other words, when $\omega$ is drawn from $p$, $\zeta_{\omega}(\mathbf{u})\zeta_{\omega}(\mathbf{v})^*$ is an unbiased estimate of $k(\mathbf{u}-\mathbf{v})$. If both the probability $p(\omega)$ and $k(\mathbf{u}-\mathbf{v})$ are real,
$\psi_{\omega}(\mathbf{u})=[\textrm{cos}(\mathbf{u})\ \textrm{sin}(\mathbf{u})]^\top$ gives a mapping so that $\mathbb{E}[\psi_{\omega}(\mathbf{u})]^\top\psi_{\omega}(\mathbf{v})=k(\mathbf{u},\mathbf{v})$. Thus, drawing $D$ i.i.d. samples $\omega_1,\ldots, \omega_D \in \reals^d$ from $p$ leads to a randomized feature map $\reals^d \mapsto \reals^{2D}$
\begin{equation}
\begin{split}
\label{Eq3}
\psi(\mathbf{u}):=\sqrt{\frac{1}{D}}[\textrm{cos}(\omega_1^{\top}\mathbf{u}),\ldots,\textrm{cos}(\omega_D^{\top}\mathbf{u}),\\
\textrm{sin}(\omega_1^{\top}\mathbf{u})
\ldots, \textrm{sin}(\omega_D^{\top}\mathbf{u})]^{\top},
\end{split}
\end{equation}
such that $\langle \psi(\mathbf{u}),\psi(\mathbf{v}) \rangle = k(\mathbf{u}-\mathbf{v})$.
\begin{figure*}
\centering
\includegraphics[scale=0.53, trim=0 100 0 0]{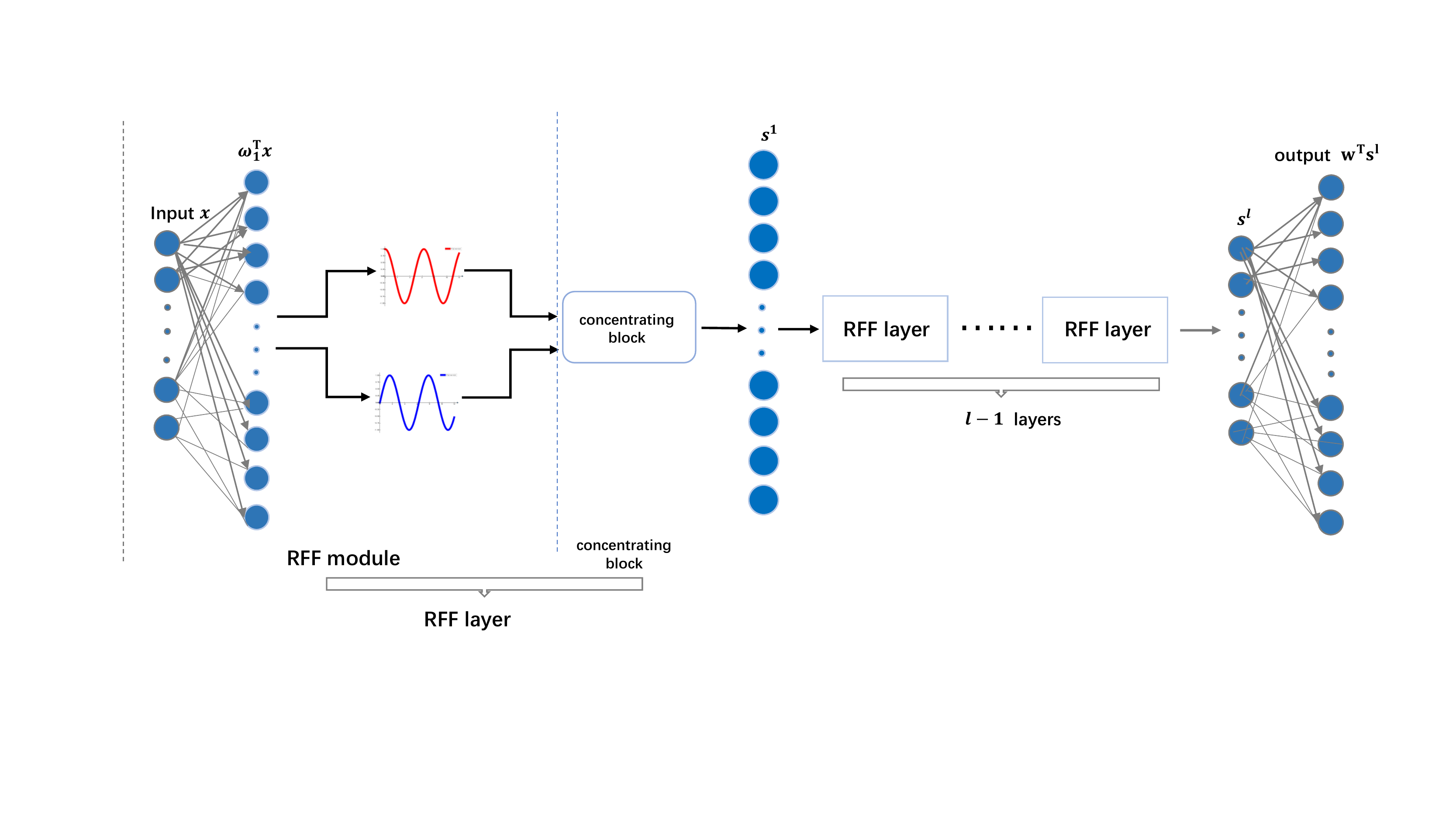}
\caption{Our deep kernel learning model via random Fourier features, illustrated between the input layer and the output layer. The data points are multiplied by corresponding $\omega$, mapped by $cos(\cdot)$ and $sin(\cdot)$ respectively and then be concatenated in the concentrating block and become the inputs of the next layer (we call the parts between the two imaginary lines a RFF module). The final layer is a fully connected layer which produces the corresponding output such as the labels in classification tasks. With supervision, the kernels in different layers are learned by back propagation. Note the dimension $D$ of $\omega$ in each RFF module could be different.  }
\label{Fig1}
\end{figure*}
Many kernels, including the RBF kernel \cite{vempati2010generalized}, the Laplacian kernel, and the Cauchy kernel, satisfy the shift-invariant and definiteness condition.
For the RBF kernel $k(\mathbf{u}-\mathbf{v})=e^{-{||\mathbf{u}-\mathbf{v}||^2_2}/{2}}$, the distribution could be explicitly given $p(\omega) = (2\pi)^{-{D}/{2}}
e^{-{||\omega||^2_2}/{2}}$. For other kernels, the distribution could be numerically calculated.
Thanks to the explicit feature map provided by RFF, one can avoid the $\mathcal{O}(n^2)$ memory and $\mathcal{O}(n^3)$ computation of the kernel matrix, which is now popular for speeding up kernel machines \cite{rahimi2008random}, \cite{yang2012nystrom}, \cite{yu2016orthogonal}.

For speeding up kernel calculation, the flowchart is to first choose an appropriate kernel and then apply RFF to approximate it. In this paper, we take the advantage of RFF for approaching kernels to incorporate deep architecture into the kernel learning. In the proposed structure, the distribution of $\omega$ in each layer could be trained and it is expected that via training the distribution we can find suitable kernel for each layer, which is at least theoretically feasible since RFF covers many kernels.

\section{Random Fourier features neural networks}

\subsection{Structure}
With random Fourier Features, we can approximate kernels with feature mappings determined by trainable distributions. Therefore, we now could realize the deep kernel structure.  Specifically, our deep kernel learning framework via random Fourier features is demonstrated in Fig. \ref{Fig1} and called random Fourier features neural networks (RFFNet). In RFFNet, there are $l$ layers, each of which consists of a RFF module and a concentrating block.
 A RFF module is the key part for producing features, including
 linear transformation, $\textrm{cos}(\cdot)$ and  $\textrm{sin}(\cdot)$ mappings. The last layer is usually a fully connected layer which carries out the linear transform and yields the corresponding
output such as the probabilities of the labels in classification tasks.

In RFF module, the input first goes through a linear transformation, where the parameters $\omega_i=[\omega^1_i,\ldots,\omega^D_i]$ are corresponding to a distribution related to a kernel and the important feature part in RFF is expressed as $\mathbf{f}=\omega_i^\top\mathbf{x}$.
Here, for vectors, this operation could be executed by a full-connection
layer without bias. For image data, considering a 3-channel image $\mathbf{X}$
of size $N_h\times N_w\times 3$ and let $\mathbf{x} \in \mathbb{R}^{N}$,
where $N=N_h\cdot N_w \cdot 3$, be the vector that is constructed by stacking
the column of $\mathbf{X}$ together. And the feature $\mathbf{f}$ could be computed by passing the image $\mathbf{X}$ from a convolutional layer.

 Then the feature $\mathbf{f}$ is mapped by the cosine, sine function and stacked in the concentrating block in order to yield the corresponding random Fourier features with regard to $\omega_i$. Besides the stacking, the concentrating block for image data could be also integrated with more
 functional structure or tool such as the batch normalization \cite{ioffe2015batch},
 the dropout \cite{srivastava2014dropout}, the residual connection \cite{he2016deep} and the pooling, etc. In the proposed model, we incorporate the batch normalization and pooling operation into the concentrating block. By reducing internal covariate shift, batch normalization
 tremendously speeds up the training process, enables higher learning rates and regularizes the model to some extent. And the RFFnet also benefits from the pooling operation due to its augmentation for the noise tolerance and reduction of the parameters to be trained.
\begin{figure*}
\centering
\includegraphics[scale=0.75, trim=0 0 00 0]{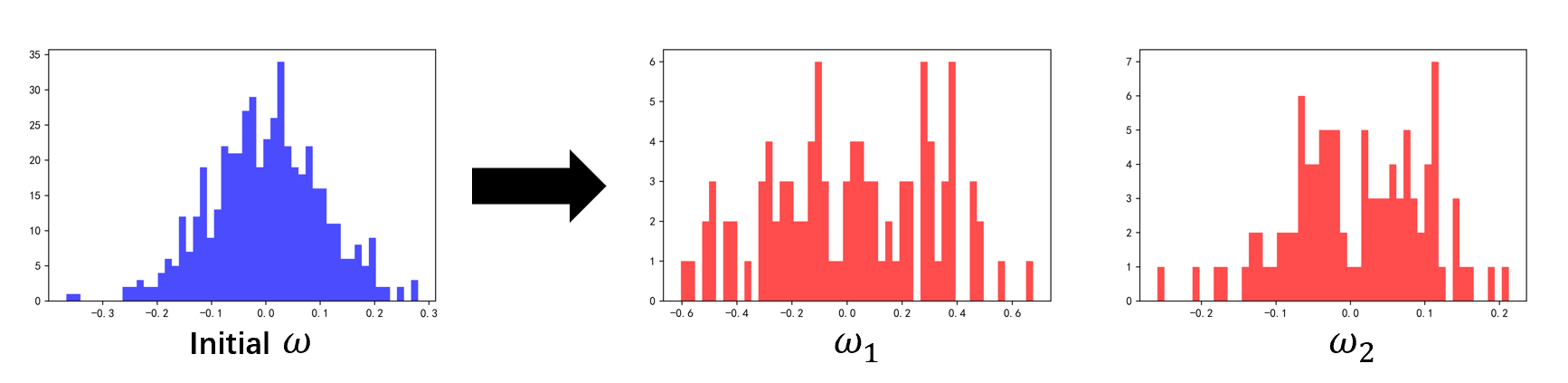}
\caption{
The histograms of $\omega$ (in the first dimension) in the two-layer model for ``monks1".
  }
\label{Fig2}
\end{figure*}

\begin{figure*}
\setlength{\belowcaptionskip}{-0.0cm}
\setlength{\abovecaptionskip}{-0.0cm}
\centering
\includegraphics[scale=0.4, trim=0 0 00 00]{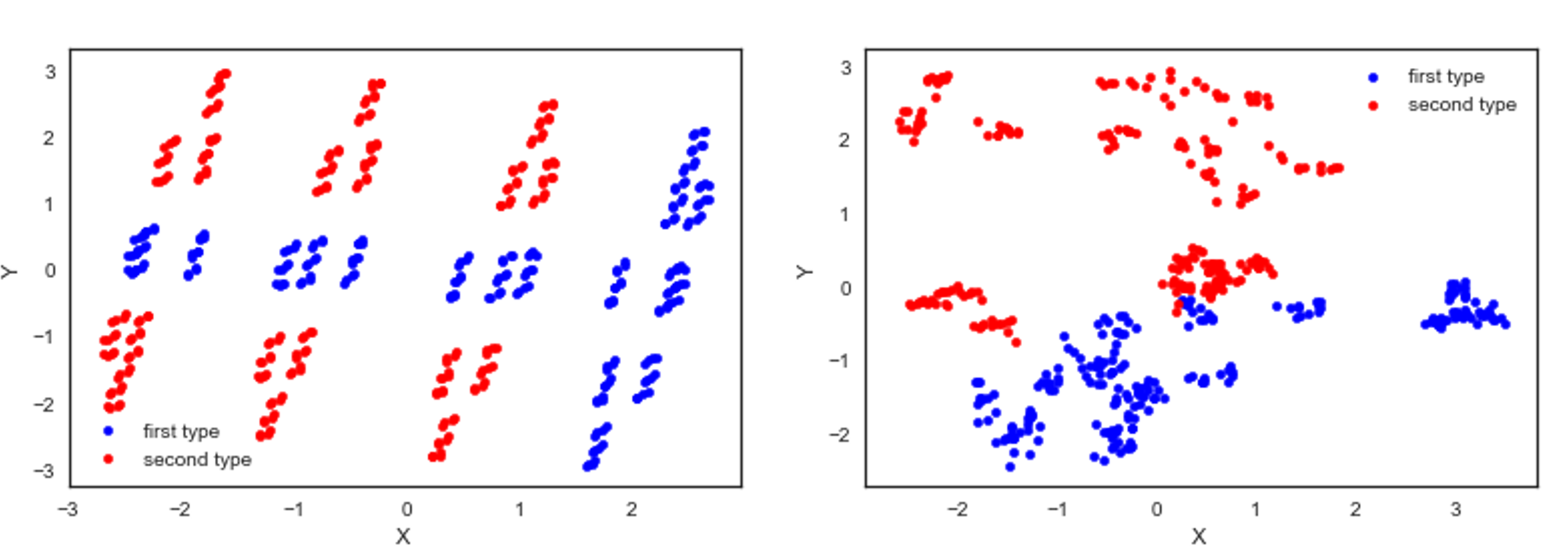}
\caption{
The visualization of random Fourier features in two layers for the classification task \textbf{monks1}. The two classes are marked by the red and blue points. From the first picture we could see the two classes in the first layer are still not linear-separable. While mapped
again by the kernel in the second layer, the two classes are nearly separated from each other, which results in a 100\% test accuracy.
  }
\label{Fig3}
\end{figure*}

Through one layer which contains of a RFF module and a concentrating block, data are projected into a finite-dimensional feature space and then could be treated as the input of the next layer. For instance, the input sample $\mathbf{x} \in \reals^d$ is transformed into the feature $\mathbf{s}_1 \in \reals^{2D}$, where $D$ is the number of RFFs. As illustrated in (\ref{rff}), with a larger $D$, the empirical average gives a better estimation to the expectation, i.e., the kernel is approximated better. By stacking layers one by one as described above, we are able to build up an end-to-end deep kernel learning architecture for different tasks.

\subsection{Training}

We apply end-to-end training to find suitable model which essentially means that we are training each layer among all the shift-invariant and positive definite kernels.
Consider training samples $\{\mathbf{x}_1,\mathbf{x}_2,\ldots,\mathbf{x_n}\}$ with labels $\{y_1,\ldots,y_n\}$, we solve the following problem to establish the discriminate function
\begin{align}
\label{Eq9}
\mathop{\textrm{min}}\limits_{f} \frac{1}{n}\sum_{i=1}^n
L(y_i,f(\mathbf{x_i}))+\frac{\lambda}{2}||f||^2,
\end{align}
where $f$ is the prediction function, $\lambda$ is the penalizing parameter to control
the complexity, and $L:\mathbb{R}\times\mathbb{R}\mapsto \mathbb{R}$ is a smooth loss function.

The training procedure follows the standard way of back-propagation. The particular operation in RFFNet is the $\sin$ and $\cos$. Representing $W=\{\omega_1,\ldots,\omega_l\}$, $\omega_i=[\omega^1_i,\ldots,\omega^{D_i}_i]$ as the parameters in
the RFF modules and $w$ as the parameters in the last linear layer,  utilizing $E_i$ for loss of $\mathbf{x_i}$
 and $\mathbf{s_j}$ for the output of the $j^{th}$ RFF layers,
 we could update $\omega_i^m$ by the chain rule. For example, we have
\begin{align}
\frac{\partial E_i}{\partial \omega_{l}^m}=
\frac{\partial E_i}{\partial \tilde{y_i}}\cdot
(
{\frac{\partial \tilde{y_i}}{\partial {s_l^m}}}
\cdot
{\frac{\partial {s_l^m}}{\partial {\omega_l^m}}}
+
{\frac{\partial \tilde{y_i}}{\partial {s_l^{m+D_l}}}}
\cdot
{\frac{\partial {s_l^{m+D_l}}}{\partial {\omega_l^m}}}
),
\end{align}
where $D_l$ is the dimension of $\omega_l$, $\tilde{y_i}=w^{\top}s_l+b$,
\begin{align}
\frac{\partial \tilde{y_i}}{\partial s_{l}^m}=w_m,\ \ \
\frac{\partial \tilde{y_i}}{\partial s_{l}^{m+D_l}}=w_{m+D_l},\\
{\frac{\partial {s_l^m}}{\partial {\omega_l^m}}}=-s_{l-1}\textrm{sin}[(\omega_l^{m})^{\top}s_{l-1}]\\
\ \ \ {\frac{\partial {s_l^{m+D_l}}}{\partial {\omega_l^m}}}=s_{l-1}\textrm{cos}[(\omega_l^{m})^{\top}s_{l-1}].
\end{align}

Since the training problem is non-convex, the initialization may have influence. In this paper, we generate initial $\omega$ from Gaussian distributions, which corresponds to using RBF kernels in each layer. RBF kernels' performance is generally not bad, as verified in many tasks, our initial setting gives a relatively good cascade structure.

\subsection{Analysis}

\textbf{Trainable:}\ \ In RFFNet, we implement a deep structure and its flexility is expected to be improved. One advantage of using RFF is its wide coverage to a rich class of kernels. If we can efficiently train the parameters, suitable kernels could be found for each layer. Let us consider an example on dataset ``monks1'', which contains 124 training samples with 6 features.  By support vector machine (SVM) with RBF kernel, of which the parameter is tuned by cross-validation, the accuracy is about 81.5\%. By setting $\omega$ from Gaussian distribution, which is actually our initialization strategy, classification by RFF has similar performance. But in our RFFNet, we can train the distribution and the obtained distribution is displayed in Fig. \ref{Fig2}, which shows that after training, the kernel has been changed to another shift-invariant and positive definite one.
\begin{figure*}[!htb]
\centering
\includegraphics[scale=0.55, trim=0 0 00 00]{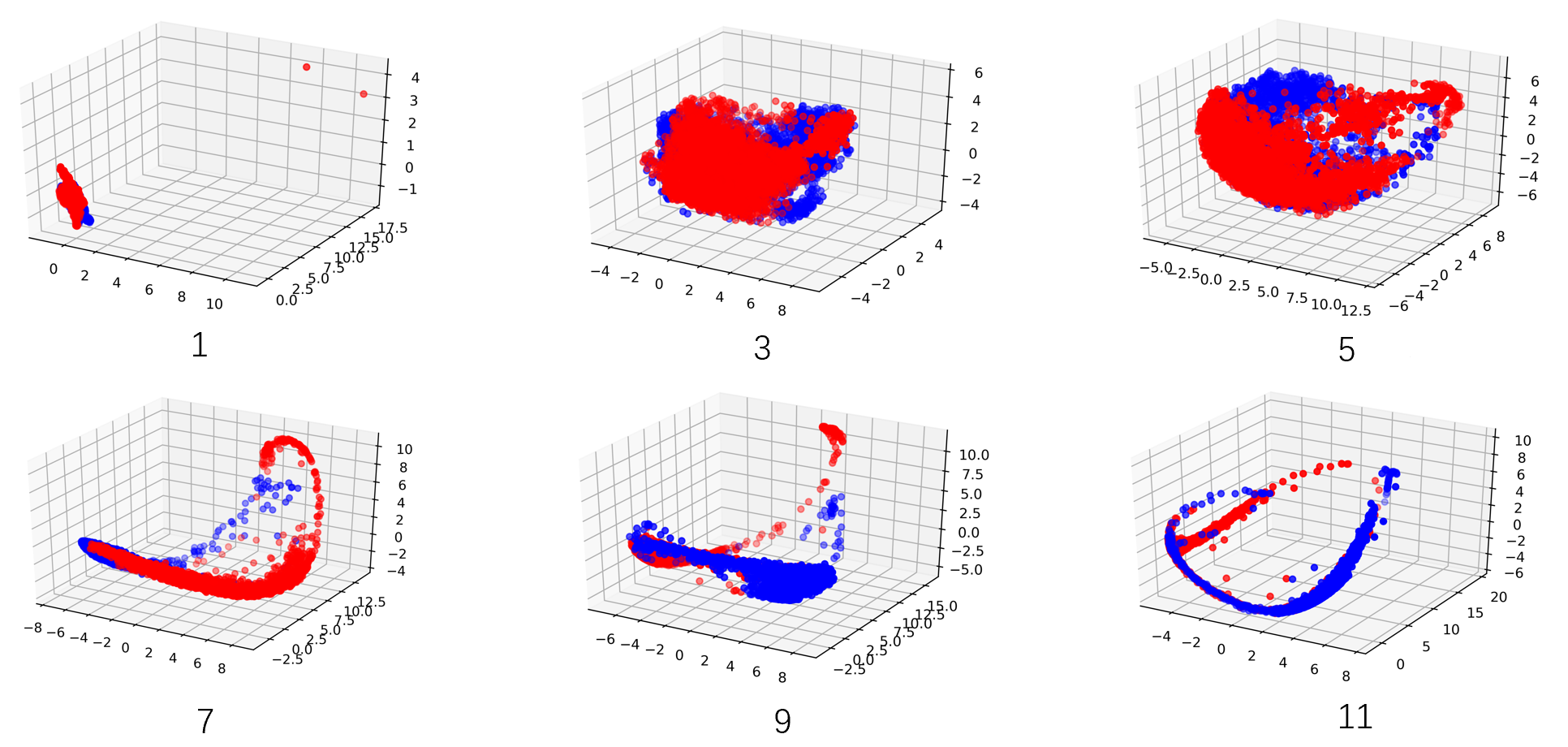}
\caption{
The visualization of random Fourier features in different layers in $\textbf{EEG}$ with a 11-layer model. The first row represents the results for layer 1, 3, 5 from the
left to right. And the second row is corresponding to layer 7,9 and 11 respectively. Being mapped by the kernels layer by layer,
the features gradually spread apart. The final test accuracy is about 98.1\%.
  }
\label{Fig4}
\end{figure*}




\textbf{Features layer by layer:} \ \ With a deep structure and efficient training, RFFNet can significantly improve the performance of shallow kernel methods, if the flexibility is the bottleneck. For example, in the mentioned ``monks1'' dataset, RFFNet with two layers has $100\%$ accuracy on the test data. The kernel method is to learn the pairwise relationship to formulate the kernel matrix. For ``monks1'', let us check the kernel matrix in each layer, by applying kernel principal component analysis (kPCA, \cite{scholkopf1998nonlinear}) to reduce features in two dimensional space. Fig.~\ref{Fig3} shows that the kernel matrix in the first layer cannot distinguish the two classes, while, in the next layer, they can be perfectly classified. In other words, good pairwise information have been extracted by RFFNet.

Similar performance is observed in ``EEG'' dataset.
There are 14980 samples with 14 features in EEG. Due to
the abundant amount of the data, this task requires stronger
learning ability. By support vector machine (SVM) with RBF
kernel, of which the parameter is tuned by cross-validation, the accuracy is only about 78.5\%, while RFFNet with eleven layers
achieves about 98.1\%. To better visualize the random Fourier features, here we compute the kernel matrix in different layers, and then apply kPCA to reduce the kernel in three-dimensional space.
As illustrated in Fig.~\ref{Fig4},
in the beginning, the features are mixed together. By going up the hierarchy, the features are mapped by multiple kernels and become
disperse from the other class, which gradually possess the  linear-separable property. This phenomenon demonstrates the effectiveness of the deep architecture when coping with abundant
and complicated data.

\textbf{pair-wise measurement:}
Always conducted by the kernel trick, pair-wise measurement is a crucial
part in kernel methods, which demonstrates the similarity or dissimilarity between two samples to some extent. The potent representation ability enables the RFFnet to extract considerably precise pair-wise information.
Fig.~\ref{Figm} indicates that in the pair-wise matrices, samples with the same labels have a larger
pair-wise value, while samples that belong
\begin{figure}
\centering
\includegraphics[scale=0.132, trim=50 70 00 00]{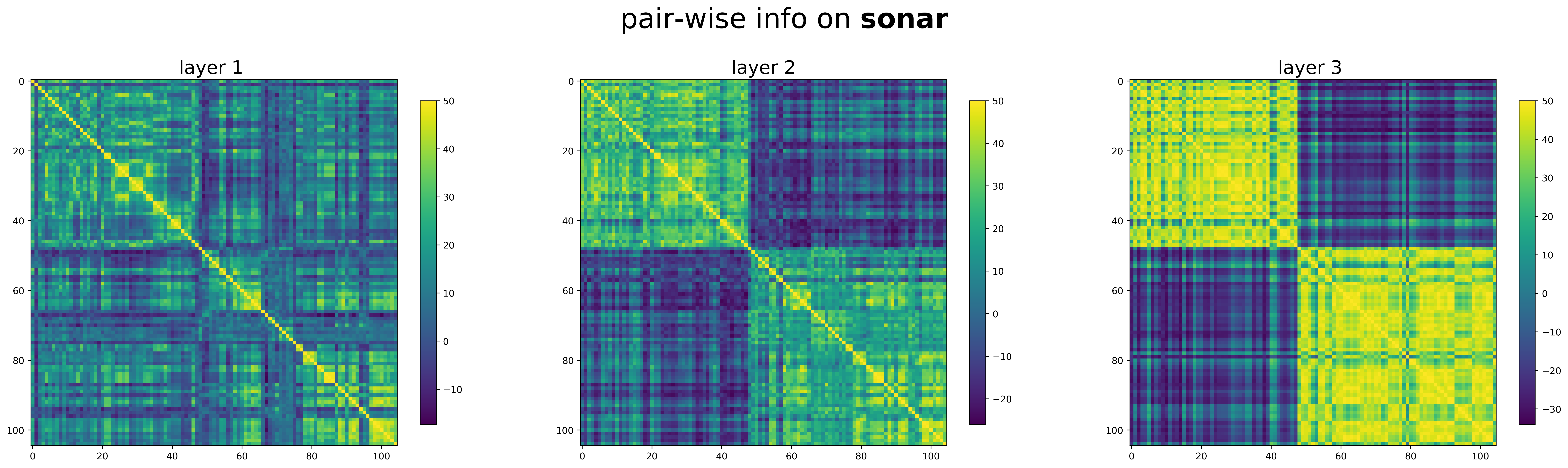}
\label{Figs}
\end{figure}
\begin{figure}
\setlength{\belowcaptionskip}{0.0cm}
\centering
\includegraphics[scale=0.132, trim=60 70 00 00]{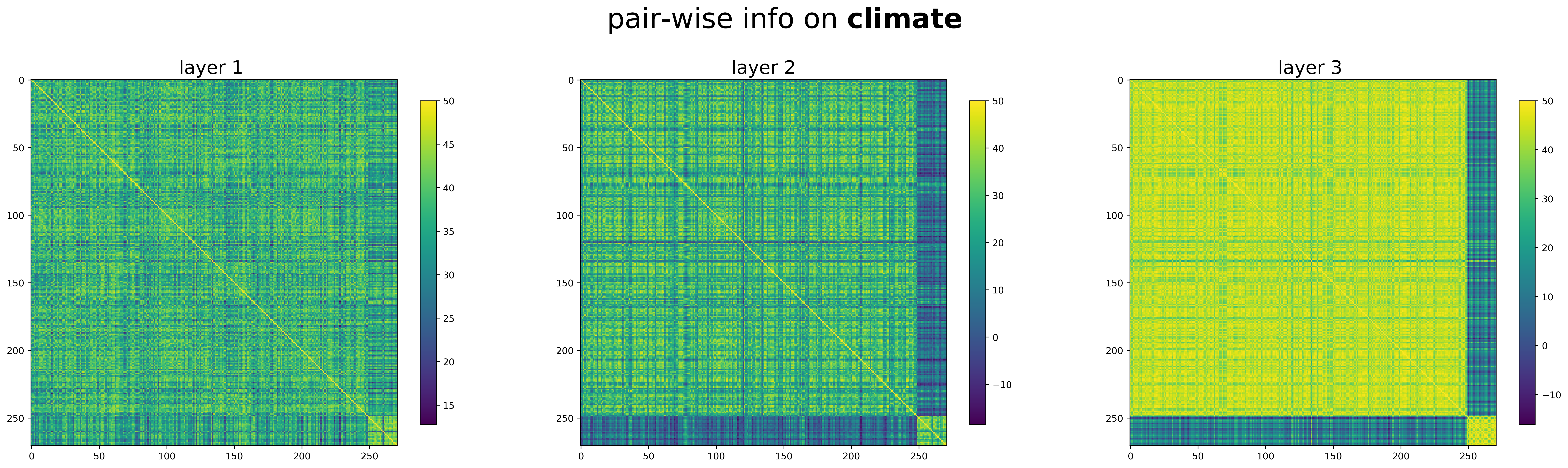}
\label{Figc}
\end{figure}
\begin{figure}
\centering
\includegraphics[scale=0.132, trim=60 50 00 00]{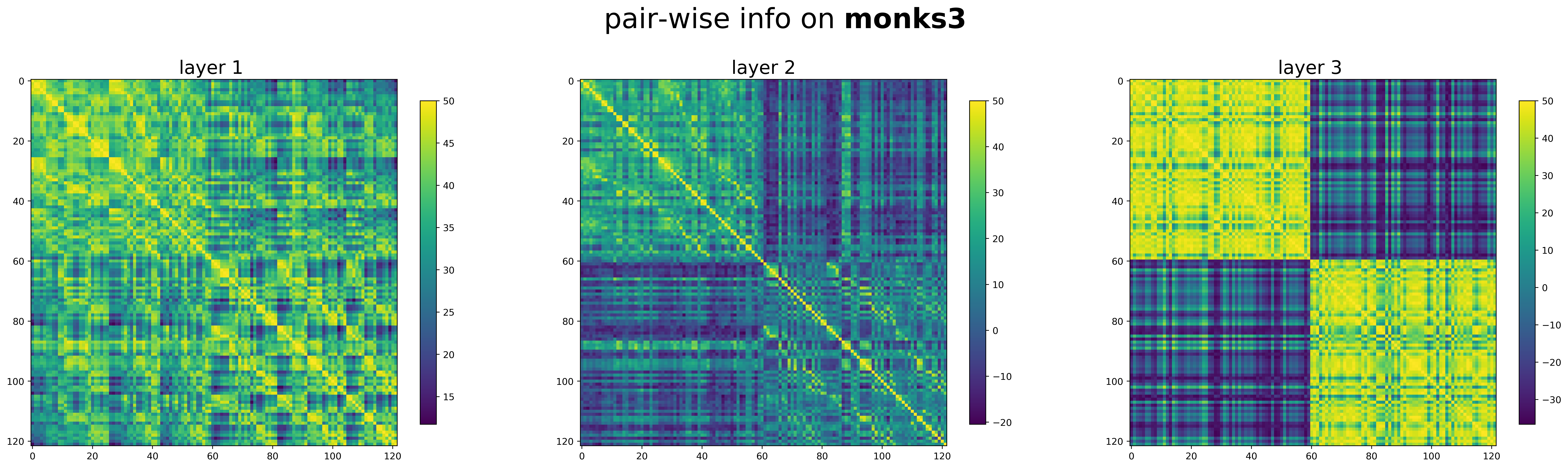}
\caption{pair-wise information on $\textbf{sonar},\textbf{climate},\textbf{monks3}$ from three-layer models.
The axises depict the  serial numbers of samples.}
\label{Figm}
\end{figure}
\begin{table*}
\centering
\caption{Average test accuracy in percents of different methods on small datasets }
\begin{tabular}{llllllll}
\toprule
Datasets
   &    size(d,n) & SVM-CV & MKL &  KRF & MLP & DKL(GP)  & RFFNet \\
  \midrule
monks1 &(6, 124) & 81.5$\pm$0.0  & 76.9$\pm$1.1 & 85.0$\pm$1.2  & 99.1$\pm$0.8   & 81.6$\pm$4.8    &  \textbf{100.0$\pm$0.0}\\
monks2 &(6, 169) & 85.8$\pm$1.0  & 63.7$\pm$5.5  & 73.9$\pm$1.3 & 84.9$\pm$2.2 & 87.8$\pm$4.9   & \textbf{98.0$\pm$0.7}\\
monks3 &(6, 122) & 93.1$\pm$1.1   & 90.1$\pm$1.2  &
 93.7$\pm$0.9 & 90.7$\pm$0.9  & 93.6$\pm$1.3
 & \textbf{93.8$\pm$0.8}\\
 fertility &(9,100) & 88.6$\pm$2.5  & 66.2$\pm$7.1 &83.0$\pm$3.7 & 82.9$\pm$3.1 & \textbf{93.2$\pm$1.8}
   & 91.2$\pm$1.8\\
  sonar&(60, 208) &
  \textbf{83.4$\pm$3.0} & 75.0$\pm$2.5 & 82.6$\pm$2.3   & 82.7$\pm$2.4 & 78.4$\pm$2.6 & 82.9$\pm$2.3\\

 climate &(20, 540) & 94.2$\pm$1.6 & 90.6$\pm$1.5 & 93.1$\pm$1.7   & 94.4$\pm$0.8 & 92.0$\pm$0.0 & \textbf{94.6$\pm$1.5}\\

 Qsar&(41,1055) & 86.7$\pm$1.3 & 84.7$\pm$1.6 & 86.9$\pm$1.6  & 84.2$\pm$0.8   & 61.6$\pm$0.5   & \textbf{86.9$\pm$0.9}\\
  \bottomrule
  \end{tabular}
  \label{tbl:table1}
\end{table*}
to different categories own a lower similarity. In the figures we are able to see obvious boundaries between two types.




\begin{table*}
\centering
\caption{Average test accuracy in percents of different methods on large scale datasets}
\begin{tabular}{lllllll}
\toprule
Datasets
   &    size(d,n) & SVM-CV &  KRF & MLP & DKL(GP)  & RFFNet \\
  \midrule
phishing &(30,11055) & 95.7$\pm$0.3 & 92.8$\pm$0.4  & 95.8$\pm$0.4  & 89.4$\pm$0.5   &  \textbf{97.1$\pm$0.2}\\
EEG &(14,14980)   & 78.5$\pm$1.2 & 55.3$\pm$0.5  & 92.5$\pm$3.4 & 61.6$\pm$2.1    & \textbf{98.1$\pm$0.4}\\
 ijcnn1 &(22,49900) & 90.7$\pm$0.0  & 93.0$\pm$0.1    & 98.1$\pm$0.1 & \textbf{98.2$\pm$0.2}
 & 97.7$\pm$0.2 \\
 covtype &(54,581012) & 80.0$\pm$0.1  & 79.0$\pm$0.2  & 96.3$\pm$0.2  & 86.1$\pm$0.1    & \textbf{96.6$\pm$0.2}\\
  \bottomrule
  \end{tabular}
  \label{tbl:table2}
\end{table*}

\section{Experiments}
In this section, comparing with other representative methods, we evaluate the proposed deep kernel learning model on a collection of several tasks, including small data problems, large-scale issues and image classification. All experiments were conducted
on a Linux machine with 12-core 3.2GHz Intel CPUs using Matlab
and PyTorch, a general deep learning platform.
For our deep kernel learning model, we utilize Adam algorithms \cite{kingma2014adam} as the optimizer with the learning rate $0.001$, coefficients betas $(0.9,0.999)$. The number of layers generally follows the rule $\lceil n/1000\rceil + 1$, where $n$ is the data size. For binary classification tasks, we use the hinge loss or the squared loss. While the cross entropy loss is exploited to tackle with multi-class classification. The input is pre-processed (normalized) with the whitening procedure. In addition, we use the multivariate Gaussian distribution $\frac{1}{(2\pi)^{\frac{n}{2}}|\Sigma|^{\frac{1}{2}}}\textrm{exp}
(-\frac{1}{2}(\omega-\mu)^{\top}\Sigma^{-1}(\omega-\mu))$to initialize $p(\omega;\mu,\Sigma)$ with
zero mean value $\mu$ and the diagonal matrix $\Sigma$ where the diagonal elements are $0.01$. This operation makes that
the initial model could be regarded as a cascade connection of multiple Guassian kernels. And the regularization parameter $\lambda$ is set to $10^{-4}$.

\begin{table*}
\centering
\caption{Test accuracy in percents on the MNIST and CIFAR-10 datasets without data augmentation }
\begin{tabular}{llllll}
\toprule
Datasets
   &     CKN\cite{mairal2014convolutional} & ResNet-18 & VGG-16 & DKL(GP) & RFFNet \\
  \midrule
 MNIST   & 99.4  & 99.3   & \textbf{99.4}   & 99.2  &  99.1\\
CIFAR-10 & 80.5   & \textbf{86.1}   & 86.0   & 77.0  &  84.6\\
  \bottomrule
  \end{tabular}
  \label{tbl:table3}
\end{table*}

\subsection{Results on small data}
When limited data are available, it is not very easy to catch
the inner characteristics. Rather than the common deep neural networks which suffer from the overfitting, kernel methods usually require less samples but have better generalization
capability. In our model, we could well approximate a kernel in each layer, hence it is quite suitable for the small data task.

To show the effectiveness, we choose some small datasets from UCI
Machine Learning Repository\footnote{http://archive.ics.uci.edu/ml/datasets.php}  \cite{blake1998uci} and compare the proposed model with
other representative method including \textbf{SVM-CV} \cite{chang2011libsvm}, $\textbf{KRF}$ \cite{sinha2016learning} \textbf{MKL}\cite{jawanpuria2015generalized}, \textbf{MLP}, and \textbf{DKL}\cite{wilson2016stochastic}. All data are normalized to [0,1]. For some problems, both the training
and the test sets are provided, otherwise we randomly pick half of
the data as the training set and the rest for test. Kernel parameters and penalty parameters in competitive approaches are tuned by 5-fold cross validation. The LIBSVM \cite{chang2011libsvm} toolbox is applied for the SVM-CV with a Gaussian kernel. The initial set in MLP follows
our model.
As illustrated in Table \ref{tbl:table1}, we report
the average test accuracy and the standard deviation on $10$ trials of different approaches.

As a deep structure, RFFNet has significant improvement from shallow kernel methods, for which
SVM-CV is a good representative. For instance, for ``monks1'', ``monks2'', and ``fertility'', the
accuracy is improved to more than 90\%, which is almost impossible for shallow methods. But of
course, if there is no problem in flexibility for shallow models, the improvement is slight.
Compared with other representative approaches, the proposed model significantly benefits from
both the flexible, learnable kernel and the deep architecture and hence owns a strong representation capability. For the dataset ``monks1'', the proposed model achieves
100\% test accuracy, which is slightly higher than 99.1\% for MLP
and much higher than that for other methods. In ``monks2'', our model outperforms over 10\% than others. Due to the limited
data, MLP has a low performance on ``fertility'', our result is 91.2\% while the DKL(GP) achieves a high performance 93.2\%. But in DKL(GP),
it actually contains a MLP and a Gaussian process layer which is more complicated than our model.
On the one hand, unlike MLP, our model takes the advantage of kernel methods and does better in alleviating the overfitting problems. On the other hand, the incorporation of deep architecture
tremendously improves the richness of kernels and thus
our model could have a satisfactory performance. In addition, we could see that MKL generally performs
not as well as other methods, hence we do not list it in the following experiments.

\subsection{Results on large scale data}
Training on a large amount of data could alleviate the problem of
overfitting while requires stronger representative capability.
To evaluate the proposed model in large scale
situations, we pick several large scale datasets ``phishing'', ``EEG'', ``ijcnn1'' and ``covtype'' from the UCI repository. In this task, we set 11 as the number of layers in our model and the same to MLP. For extremely large datasets like
``covtype'', we train the model with a minibatch size of 256 and the maximum epochs are 300.

As shown in Table \ref{tbl:table2}, we achieve promising performance on these large scale datasets, which ranks first
 in 3 datasets. Due to the deep
structure, the learning ability of our model and MLP
are remarkably stronger than
SVM-CV and other similar kernel-based approaches, which are unable to perform as well as our model and MLP. In ``EEG'', the ``shallow'' kernel methods, including SVM and KRF, do not have a
satisfactory accuracy. Especially the KRF, which also has a supervised approach to learn a kernel, only gets 55.3\%. As for DKL(GP), it is possible that the feature extraction layer in DKL(GP) could not catch the valuable information in the samples of ``EEG'' hence it performs not so well.

For the large scale data, broadly speaking, the models equipped with deep architecture possess better learning ability and strong
representation capability than shallow ones. Rather than MLP with rectified linear units, the proposed RFFNet owns a powerful kernel in each layer, which leads to a strong non-linearity
and the richness to describe the inner distribution of the data.

\subsection{Image data MNIST and CIFAR-10}
We now move to the image datasets MNIST\footnote{http://yann.lecun.com/exdb/mnist/} \cite{lecun1998gradient} and CIFAR-10\footnote{http://www.cs.toronto.edu/~kriz/cifar.html} \cite{krizhevsky2012imagenet}.  The MNIST dataset consists of handwritten digit images and it is divided in 60,000 examples for the training set and 10,000 examples for testing. CIFAR-10 is an established computer-vision
dataset used for object recognition, which
has 60000 32$\times$32 color images with 10 classes.

Here we evaluate the performance with  a
9-layer model, which has the batch normalization and 3 max-pooling layers for (2, 2), (2, 2), (8, 8), respectively. In addition, here we use $3\times3$ as convolutional kernel size, the stride is one with one padding. The transformation of feature maps is mainly 3-64-64-128-128-256-256. The number of epochs is at most 300 with a minibatch size of 256. Beginning from an
initial learning rate of 0.01, we drop it by a factor of 10 after 200 epochs and then again after 240
epochs. Other methods are equipped with their recommended settings.

We present the experimental results in Table \ref{tbl:table3} with
the performance of different approaches without data augmentation.
Compared with other kernel-based approaches, the results of VGG-16 \cite{simonyan2014very} and ResNet-18 \cite{he2016deep} are outstanding. But our model
 also could achieve reasonably comparable performance with 99.1\%
 on MNIST and 84.6\% on CIFAR-10. Notice that RFFNet
has not been specifically designed and polished for image processing. With suitable supplementary
blocks and techniques, the performance may have improvement space.

\section{Conclusion}
In this paper, we propose a deep structure for kernel learning by random Fourier features. On the on hand, RFF provides a good approximation for kernel, which leads to good performance in small datasets. On the other hand, RFF makes it possible to construct multi-layer networks and the kernel in each layer could be efficiently trained by back propagation. Without tricky designs, the proposed RFFNet achieves good performance on small data, large scale data and image data such as MNIST and CIFAR-10. Generally, RFFNet is a promising direction to make kernel methods deeper and it may also bridge the theoretical analysis for shallow kernel methods and deep learning.

\bibliographystyle{aaai}
\bibliography{nips5}

\begin{thebibliography}{}

\bibitem[\protect\citeauthoryear{Blake}{1998}]{blake1998uci}
Blake, C.
\newblock 1998.
\newblock \protect{UCI} repository of machine learning databases.
\newblock {\em http://www. ics. uci. edu/\~{} mlearn/MLRepository. html}.

\bibitem[\protect\citeauthoryear{Chang and Lin}{2011}]{chang2011libsvm}
Chang, C.-C., and Lin, C.-J.
\newblock 2011.
\newblock {LIBSVM}: A library for support vector machines.
\newblock {\em ACM Transactions on Intelligent Systems and Technology (TIST)}
  2(3):27.

\bibitem[\protect\citeauthoryear{Cho and Saul}{2009}]{cho2009kernel}
Cho, Y., and Saul, L.~K.
\newblock 2009.
\newblock Kernel methods for deep learning.
\newblock In {\em NIPS}.

\bibitem[\protect\citeauthoryear{He \bgroup et al\mbox.\egroup
  }{2016}]{he2016deep}
He, K.; Zhang, X.; Ren, S.; and Sun, J.
\newblock 2016.
\newblock Deep residual learning for image recognition.
\newblock In {\em CVPR}.

\bibitem[\protect\citeauthoryear{Ioffe and Szegedy}{2015}]{ioffe2015batch}
Ioffe, S., and Szegedy, C.
\newblock 2015.
\newblock Batch normalization: Accelerating deep network training by reducing
  internal covariate shift.
\newblock {\em arXiv preprint arXiv:1502.03167}.

\bibitem[\protect\citeauthoryear{Jawanpuria, Nath, and
  Ramakrishnan}{2015}]{jawanpuria2015generalized}
Jawanpuria, P.; Nath, J.~S.; and Ramakrishnan, G.
\newblock 2015.
\newblock Generalized hierarchical kernel learning.
\newblock {\em Journal of Machine Learning Research} 16(1):617--652.

\bibitem[\protect\citeauthoryear{Kingma and Ba}{2015}]{kingma2014adam}
Kingma, D.~P., and Ba, J.
\newblock 2015.
\newblock Adam: A method for stochastic optimization.
\newblock In {\em ICLR}.

\bibitem[\protect\citeauthoryear{Krizhevsky, Sutskever, and
  Hinton}{2012}]{krizhevsky2012imagenet}
Krizhevsky, A.; Sutskever, I.; and Hinton, G.~E.
\newblock 2012.
\newblock Image{N}et classification with deep convolutional neural networks.
\newblock In {\em NIPS}.

\bibitem[\protect\citeauthoryear{LeCun \bgroup et al\mbox.\egroup
  }{1998}]{lecun1998gradient}
LeCun, Y.; Bottou, L.; Bengio, Y.; Haffner, P.; et~al.
\newblock 1998.
\newblock Gradient-based learning applied to document recognition.
\newblock {\em Proceedings of the IEEE} 86(11):2278--2324.

\bibitem[\protect\citeauthoryear{Mairal \bgroup et al\mbox.\egroup
  }{2014}]{mairal2014convolutional}
Mairal, J.; Koniusz, P.; Harchaoui, Z.; and Schmid, C.
\newblock 2014.
\newblock Convolutional kernel networks.
\newblock In {\em NIPS}.

\bibitem[\protect\citeauthoryear{Mairal}{2016}]{mairal2016end}
Mairal, J.
\newblock 2016.
\newblock End-to-end kernel learning with supervised convolutional kernel
  networks.
\newblock In {\em NIPS}.

\bibitem[\protect\citeauthoryear{Rahimi and Recht}{2008}]{rahimi2008random}
Rahimi, A., and Recht, B.
\newblock 2008.
\newblock Random features for large-scale kernel machines.
\newblock In {\em NIPS}.

\bibitem[\protect\citeauthoryear{Rudin}{1962}]{rudin1962fourier}
Rudin, W.
\newblock 1962.
\newblock {\em Fourier analysis on groups}, volume 121967.
\newblock Wiley Online Library.

\bibitem[\protect\citeauthoryear{Sch{\"o}lkopf and
  Smola}{2001}]{scholkopf2001learning}
Sch{\"o}lkopf, B., and Smola, A.~J.
\newblock 2001.
\newblock {\em Learning with kernels: support vector machines, regularization,
  optimization, and beyond}.
\newblock MIT press.

\bibitem[\protect\citeauthoryear{Sch{\"o}lkopf, Smola, and
  M{\"u}ller}{1998}]{scholkopf1998nonlinear}
Sch{\"o}lkopf, B.; Smola, A.; and M{\"u}ller, K.-R.
\newblock 1998.
\newblock Nonlinear component analysis as a kernel eigenvalue problem.
\newblock {\em Neural Computation} 10:1299--1319.

\bibitem[\protect\citeauthoryear{Simonyan and
  Zisserman}{2015}]{simonyan2014very}
Simonyan, K., and Zisserman, A.
\newblock 2015.
\newblock Very deep convolutional networks for large-scale image recognition.
\newblock In {\em ICLR}.

\bibitem[\protect\citeauthoryear{Sinha and Duchi}{2016}]{sinha2016learning}
Sinha, A., and Duchi, J.~C.
\newblock 2016.
\newblock Learning kernels with random features.
\newblock In {\em NIPS}.

\bibitem[\protect\citeauthoryear{Srivastava \bgroup et al\mbox.\egroup
  }{2014}]{srivastava2014dropout}
Srivastava, N.; Hinton, G.; Krizhevsky, A.; Sutskever, I.; and Salakhutdinov,
  R.
\newblock 2014.
\newblock Dropout: a simple way to prevent neural networks from overfitting.
\newblock {\em Journal of Machine Learning Research} 15(1):1929--1958.

\bibitem[\protect\citeauthoryear{Vempati \bgroup et al\mbox.\egroup
  }{2010}]{vempati2010generalized}
Vempati, S.; Vedaldi, A.; Zisserman, A.; and Jawahar, C.
\newblock 2010.
\newblock Generalized rbf feature maps for efficient detection.
\newblock In {\em BMVC}.

\bibitem[\protect\citeauthoryear{Wilson \bgroup et al\mbox.\egroup
  }{2016a}]{wilson2016stochastic}
Wilson, A.~G.; Hu, Z.; Salakhutdinov, R.~R.; and Xing, E.~P.
\newblock 2016a.
\newblock Stochastic variational deep kernel learning.
\newblock In {\em NIPS}.

\bibitem[\protect\citeauthoryear{Wilson \bgroup et al\mbox.\egroup
  }{2016b}]{wilson2016deep}
Wilson, A.~G.; Hu, Z.; Salakhutdinov, R.; and Xing, E.~P.
\newblock 2016b.
\newblock Deep kernel learning.
\newblock In {\em AISTATS}.

\bibitem[\protect\citeauthoryear{Yang \bgroup et al\mbox.\egroup
  }{2012}]{yang2012nystrom}
Yang, T.; Li, Y.-F.; Mahdavi, M.; Jin, R.; and Zhou, Z.-H.
\newblock 2012.
\newblock Nystr{\"o}m method vs random {F}ourier features: A theoretical and
  empirical comparison.
\newblock In {\em NIPS}.

\bibitem[\protect\citeauthoryear{Yu \bgroup et al\mbox.\egroup
  }{2016}]{yu2016orthogonal}
Yu, F. X.~X.; Suresh, A.~T.; Choromanski, K.~M.; Holtmann-Rice, D.~N.; and
  Kumar, S.
\newblock 2016.
\newblock Orthogonal random features.
\newblock In {\em NIPS}.

\end{thebibliography}

\end{document}